\documentclass{article}

\usepackage{natbib}
\setcitestyle{authoryear,open={(},close={)}} 
\usepackage[preprint]{ewrl_2023}

\usepackage[utf8]{inputenc} 
\usepackage[T1]{fontenc}    
\usepackage{hyperref}       
\usepackage{url}            
\usepackage{booktabs}       
\usepackage{amsfonts}       
\usepackage{nicefrac}       
\usepackage{microtype}      
\usepackage{xcolor}         

\usepackage{amsmath}
\usepackage{amssymb}
\usepackage{mathtools}

\usepackage{lipsum}

\title{Physics-informed reinforcement learning via probabilistic co-adjustment functions}

\author{%
  Nat Wannawas\\
  Department of Bioengineering\\
  Imperial College London\\
  London, UK, SW6 3LP \\
  \texttt{nat.wannawas18@imperial.ac.uk} \\
  \And
  A. Aldo Faisal \\
  Department of Bioengineering and of Computing \\
  Imperial College London \\
  London, UK, SW6 3LP \\
  \texttt{a.faisal@imperial.ac.uk} \\
}

\begin{document}

\maketitle

\begin{abstract}
Reinforcement learning of real-world tasks is very data inefficient, and extensive simulation-based modelling has become the dominant approach for training systems. However, in human-robot interaction and many other real-world settings, there is no appropriate one-model-for-all due to differences in individual instances of the system (e.g. different people) or necessary oversimplifications in the simulation models. This requires two approaches: 1. either learning the individual system's dynamics approximately from data which requires data-intensive training or 2. using a complete digital twin of the instances, which may not be realisable in many cases. We introduce two approaches: co-kriging adjustments (CKA) and ridge regression adjustment (RRA) as novel ways to combine the advantages of both approaches. Our adjustment methods are based on an auto-regressive AR1 co-kriging model that we integrate with GP priors. This yield a data- and simulation-efficient way of using simplistic simulation models (e.g., simple two-link model) and rapidly adapting them to individual instances (e.g., biomechanics of individual people). Using CKA and RRA, we obtain more accurate uncertainty quantification of the entire system's dynamics than pure GP-based and AR1 methods. We demonstrate the efficiency of co-kriging adjustment with an interpretable reinforcement learning control example, learning to  control a biomechanical human arm using only a two-link arm simulation model (offline part) and CKA derived from a small amount of interaction data (on-the-fly online). Our method unlocks an efficient and uncertainty-aware way to implement reinforcement learning methods in real world complex systems for which only imperfect simulation models exist.
\end{abstract}

\section{Introduction}

Reinforcement Learning learns to control a physical system via trial and error and its ability to learn complex controls without requiring hand-crafted rules offers appealing in manipulation tasks, including robotics \citep{Kormushev2013,Nguyen2019}. However, conventional reinforcement learning is a data-intensive process that requires a large number of interactions. Generating a large amount of training data in real-time is challenging for human-robot interaction tasks. One dominant solution to reduce the number of interactions with the real system is to use a simulated environment to improve the sample efficiency \citep{Zhao2021}. This approach is generally referred as model-based reinforcement learning \citep{Sutton1990}.

One common approach to building the simulated environments is to use function approximators to learn the transition dynamics from the real interaction data. In recent years, the merger of reinforcement learning with the different function approximators has been observed to address high-dimensional robotic manipulations \citep{ramesh2022physics,Nagabandi2017, Deisenroth2011,andersson2015model}. One such model-based reinforcement learning algorithm is hybrid model-based model-free \citep{Nagabandi2017} that uses neural networks as function approximators and simulates different mobile robot agents. In other works by Deisenroth et al. \citep{Deisenroth2011} and Andersson et al. \citep{andersson2015model}, Gaussian Process (GP), a gold standard in terms of sample efficiency and uncertainty quantification, is used as a function approximator with reinforcement learning to simulate the models of cart-pole and cart-double pendulum and extended cart-pole, respectively. Since these modelling approaches do not require prior knowledge about the underlying robot dynamics, therefore, are easy to implement. However, it may require a substantial amount of interaction data to learn accurate robot models. Another approach is to build simulated environments using the known physics of the actual settings. An example of such an approach appears in \citep{Wannawas2021,Wannawas2022}, which uses biomechanical simulation software to create simulated human body environments. Even though this approach does not require the interaction data, the physics-derived model is not guaranteed to be accurate, especially when the environments are complex.

There are approaches in machine learning that combine the data and physical dynamics in the modelling process. An approach that received lots of attention recently is physics-informed neural networks (PINNs) \citep{RAISSI2019}, which utilise the derivative loss computed from the physics knowledge to update the network's parameters and learn the dynamics using a small amount of data.  Recently, PINNs have been used to estimate the motion of human musculoskeletal systems and artificial muscle actuators \citep{taneja2022feature, sun2022physics}. The physics knowledge, such as the equations of motion for many mechanical systems, are time-derivative and require grey box modelling \citep{Cen2011, Tulleken1993} where data is exploited to optimise the physics models' parameters. However, it relies heavily on the structure of the physics model: too complex a model is challenging to optimise, while an over-simplified model cannot accurately capture the dynamics of the existing system. 

In many real-world problems, we usually have the crude models of the environments or intuitive ideas about their behaviours. In a robot arm manipulation task, for example, we know that if we apply positive torque to a joint, the joint angle should change in a positive direction even though we may be unable to predict the exact amount of change. Leveraging such basic knowledge in modelling could improve the sample efficiency of the model, give sensible predictions when the data is scarce, and provide flexibility in using the inaccurate model. This leads us to multi-fidelity modelling, which leverages low-fidelity data (computationally cheap but inaccurate) and high-fidelity data (computationally expensive but accurate) to maximise the accuracy of model estimates. Specifically, we are interested in co-kriging technique \citep{Gratiet2014,Kennedy2000}, also known as the multi-fidelity Gaussian Process (GP) \citep{BREVAULT2020,Raissi16}. This technique uses two GPs to fuse the data from two fidelity levels to perform the prediction on a high fidelity level. The desirable properties of co-kriging over the other multi-fidelity modelling techniques are that it performs well in low data conditions and produces uncertainty estimate. We can consider the high- and low-fidelity data as the data from the real system and the model derived from physics principles, respectively. However, the existing multi-fidelity GP techniques, specifically AR1 co-kriging \citep{Gratiet2014,Raissi16}, has challenges when applied to reinforcement learning.

Here, we present two formulations based on classic AR1 co-kriging model \citep{Gratiet2014,Raissi16} that lends itself to the reinforcement learning setting. Our formulations, called the co-kriging adjustment (CKA) and ridge regression adjustment (RRA), can conveniently handle the tasks in high dimensional systems and produce better uncertainty estimates than simple GP-based and AR1 co-kriging methods. We demonstrate their behaviours in a benchmark task as well as their effectiveness in two reinforcement learning settings. We show that CKA can significantly improve the sample efficiency even when the physics model is overly simplified.

\section{Methods}
\label{headings}
The overview of physics-informed modelling is described as follows. Suppose we have a physics-derived function that predicts the transition of a system, $\mathbf{s}_{t+1}=f_p(\mathbf{x}_t)=f_p(\mathbf{s}_t,\mathbf{a}_t)$, where $\mathbf{s}_t$ and $\mathbf{a}_t$ is the state and action (control input) applied to the system, respectively. This function, which may not fully capture the system's dynamics, can be viewed as the crude function of the system's transition. The modelling task is to learn a function $f_a$ that adjusts $f_p$'s outputs to predict the real system's transitions expressed as $\mathbf{s}_{t+1}=f_a(f_p(\mathbf{x}_t),\mathbf{x}_t)$, using the transition data collected from the real system.

Our base idea is to build a probabilistic model that relies on $f_p$ and is aware of the uncertainty if the data are absent. When the data are present, the predictive means become close to the data, and the predictive variances are small. These behaviours are intuitive and essential for preventing overconfident issues in reinforcement learning. We present two flavours of physics-informed modelling with adjustment functions. co-kriging adjustment (CKA) and ridge regression adjustment (RRA).

\subsection{Co-kriging adjustment (CKA)}

\subsubsection{Co-kriging adjustment (CKA): formulation}
The mathematical formulation of the base idea is as follows. We design $f_p(x)$ to be adjusted by a scaling $\rho(x)$ and a bias $\delta(x)$ functions, expressed as:
\begin{equation}
\label{eq:eq1}
f_a(f_p(x),x) = \rho(x)f_p(x) + \delta(x).
\end{equation}
We make $f_a$ a probabilistic function by putting Gaussian Process priors over $\rho(x)$ and $\delta(x)$. As we want the predictive mean to be close to $f_p(x)$ in the area where the data are not available, we, therefore, set the prior means of $\rho(x)$ and $\delta(x)$ to be $1$ and $0$:
\begin{align*}
\rho(x)&\sim\mathcal{GP}(1,k_\rho(\boldsymbol{x},\boldsymbol{x'};\boldsymbol{\theta}_\rho)),\\
\delta(x)&\sim\mathcal{GP}(0,k_\delta(\boldsymbol{x},\boldsymbol{x'};\boldsymbol{\theta}_\delta)),
\end{align*}
where $k_\rho$ and $k_\delta$ are covariance functions with $\boldsymbol{\theta}_\rho$ and $\boldsymbol{\theta}_\delta$ parameters. Hence, the prior of $f_a$ becomes
\begin{equation}
f_a\sim\mathcal{GP}(f_p(\boldsymbol{x}),F_pk_\rho(\boldsymbol{x},\boldsymbol{x'})F_p^T + k_\delta(\boldsymbol{x},\boldsymbol{x'})),
\end{equation}
where $F_p$ is the diagonal matrix of $f_p(x)$.

\subsubsection{Co-kriging adjustment (CKA): learning \& prediction}
The learning of CKA is the optimisation of the covariance function parameters, $\boldsymbol{\theta}_\rho$ and $\boldsymbol{\theta}_\delta$. The optimisation follows the standard procedure for Gaussian Process briefly described as follows. We model the observations $\boldsymbol{y}$ at input locations $\boldsymbol{x}$ as the outputs of the real system with Gaussian noises:
\begin{equation}
\boldsymbol{y} = f_a(\boldsymbol{x}) + \boldsymbol{\epsilon},\;\;\;\boldsymbol{\epsilon}\sim \mathcal{N}(0,\sigma^2\boldsymbol{I}),
\end{equation}
where $\sigma$ is a Gaussian noises parameter. The distribution of $\boldsymbol{y}$ can then be expressed as:
\begin{equation}
\label{eq:ydis}
\boldsymbol{y}\sim \mathcal{N}({f_p}(\boldsymbol{x}), \underbrace{F_pk_\rho(\boldsymbol{x},\boldsymbol{x'})F_p^T+k_\delta(\boldsymbol{x},\boldsymbol{x'})+\sigma^2\boldsymbol{I}}_{\boldsymbol{K}}),
\end{equation}
The parameters $\theta_\rho$, $\theta_\delta$, $\sigma$ can be optimised by performing gradient descent to minimise the negative log marginal likelihood (NLML) computed as
\begin{equation}
\label{eq:NLML}
\begin{aligned}
NLML=&\frac{1}{2}(\boldsymbol{y}-f_p(\boldsymbol{x}))^T\boldsymbol{K}^{-1}(\boldsymbol{y}-f_p(\boldsymbol{x}))\\&+\frac{1}{2}log|\boldsymbol{K}|+\frac{N}{2}log(2\pi),
\end{aligned}
\end{equation}
where $N$ is the number of training data, the observations collected from the real system.

The prediction of CKA also follows the standard procedure of Gaussian process where the predictive means and variances at location $\boldsymbol{x_*}$ can be computed as:
\begin{align}
\label{eq:CKPImean}
\boldsymbol{\mu_{x_*}} &= f_p(\boldsymbol{x_*})+\boldsymbol{q}^T\boldsymbol{K}^{-1}(\boldsymbol{y}-f_p(\boldsymbol{x})),\\
\label{eq:CKPIvar}
\boldsymbol{var_{x_*}} &= \boldsymbol{K_{**}}-\boldsymbol{q}^T\boldsymbol{K}^{-1}\boldsymbol{q}+\sigma^2\boldsymbol{I},
\end{align}
where $\boldsymbol{K_{**}}$ and $\boldsymbol{q}$ are computed as
\begin{align*}
\boldsymbol{K_{**}}&=F_p(\boldsymbol{x_*})k_\rho(\boldsymbol{x_*,x_*'})F_p(\boldsymbol{x_*})^T+k_\delta(\boldsymbol{x_*,x_*'}),\\
\boldsymbol{q}&=F_p(\boldsymbol{x})k_\rho(\boldsymbol{x,x_*})F_p(\boldsymbol{x_*})^T+k_\delta(\boldsymbol{x,x_*}),
\end{align*}
where $F_p(\boldsymbol{x_*})$ is a diagonal matrix of $f_p(\boldsymbol{x_*})$.

\subsubsection{Relationship between CKA and AR1 co-kriging}
\label{subsec:cka}
It is worth mentioning Auto-Regressive (AR1) co-kriging, a classic method for multi-fidelity modelling \citep{BREVAULT2020,Raissi16}, to highlight its limitations and key differences compared to the proposed CKA. AR1 co-kriging models a linear relationship between two fidelity levels as
\begin{equation}
\label{eq:ar1}
f_2(x) = \rho f_1(x) + \delta(x)
\end{equation}
where $f_1(x)$ and $f_2(x)$ are the prediction at lower and higher levels; $\rho$ is a scaling factor; and $\delta(x)$ is an additive bias. In this context, $f_1(x)$ and $f_2(x)$ can be viewed as $f_p(x)$ and $f_a(x)$, respectively.

The key differences to CKA are that 1) AR1 models the scaling factor $\rho$ as either a constant or a deterministic function and that 2) AR1 uses a Gaussian process to model $f_p(x)$ instead of using $f_p(x)$ directly. This AR1 formulation has the following issues in reinforcement learning applications. The first issue is that applying a constant or a deterministic scaling to $f_1(x)$ causes the adjustment to be applied throughout the state-action space, including the locations where the real transitions are not yet observed. This means the predictive means in those locations can substantially differ from the physics-derived model without any evidential observation. The second issue is that using a Gaussian process to model $f_p(x)$ requires several data points generated from $f_p(x)$ to cover the entire state-action space. Given that AR1 has the $\boldsymbol{K}$ matrix with the size of $N_1+N_2\;\times\;N_1+N_2$, where $N_1$ and $N_2$ are the numbers of generated and observed data, it scales poorly with the dimension. In contrast, CKA's $\boldsymbol{K}$ matrix has the size of just $N_2\;\times\;N_2$, which is easier to invert. Besides the dimensional scaling issue, covering the entire state-action space with the generated data can lead to overconfident predictions.

\subsection{Ridge regression adjustment (RRA)}
The second probabilistic adjustment function presented here is based on ridge regression, a linear regression model whose coefficients are estimated by a ridge estimator. Ridge estimation is carried out on the linear regression model:
\begin{equation}
\boldsymbol{y} = \boldsymbol{X}\boldsymbol{\beta} + \epsilon,
\end{equation}
where $\boldsymbol{y}$, $\boldsymbol{X}$ and $\boldsymbol{\epsilon}$ are an observation vector, an input matrix, and a noise vector, respectively. $\boldsymbol{\beta}$ is the vector of regression coefficients whose posterior mean $\hat{\boldsymbol{\beta}}$ w.r.t. a least square cost function can be computed in closed-form:
\begin{equation}
\label{eq:ridge_closed_form_solution}
\hat{\boldsymbol{\beta}} = (\boldsymbol{X}^T\boldsymbol{X}+\lambda\boldsymbol{I})^{-1}\boldsymbol{X}^T\boldsymbol{y},
\end{equation}
where $\lambda$ is a constant penalising large regression coefficients.

An extension of ridge regression called kernel ridge regression applies a kernel $\varphi$ to transform the inputs from the original space into a feature space: $\boldsymbol{x}_i\rightarrow\boldsymbol{\Phi}_i=\varphi(\boldsymbol{x}_i)$. The regression is then carried out on the feature space:
\begin{equation}
\label{eq:kernel_ridge_closed_form_solution}
\hat{\boldsymbol{\beta}} = (\boldsymbol{\Phi}_X^T\boldsymbol{\Phi}_X+\lambda\boldsymbol{I})^{-1}\boldsymbol{\Phi}_X^T\boldsymbol{y},
\end{equation}
where $\boldsymbol{\Phi}_X$ is the feature matrix of the training inputs.

The posterior distribution of $\boldsymbol{\beta}$ can be obtained through Laplace approximation that provides a Gaussian approximation to the posterior around $\hat{\boldsymbol{\beta}}$:
\begin{equation}
\label{eq:beta_dist}
\boldsymbol{\beta} \sim \mathcal{N}(\hat{\boldsymbol{\beta}},\hat{\Sigma}=\hat{H}^{-1}),
\end{equation}
where $\hat{H}$ is a Hessian matrix computed as $\hat{H}=\boldsymbol{\Phi}_X\boldsymbol{\Phi}_X^T + \boldsymbol{I}$.

The predictive mean $\boldsymbol{\mu}_{x_*}$ and variance $\boldsymbol{var}_{x_*}$ at location $x_*$ are computed as
\begin{align}
\label{eq:RRA_predictive_mean}
\boldsymbol{\mu}_{x_*} &= \Phi_{x_*}\hat{\boldsymbol{\beta}},\\
\label{eq:RRA_predictive_var}
\boldsymbol{var}_{x_*} &= \boldsymbol{\Phi}_{x_*}^T\hat{\Sigma}\boldsymbol{\Phi}_{x_*},
\end{align}
where $\boldsymbol{\Phi}_{x_*}$ is the feature vector of $x_*$.

\subsubsection{Ridge regression adjuster (RRA): formulation}
Here, we introduce the formation of probabilistic adjustment functions based on kernel ridge regression. The base function is similar to that of CKA (Eq.\ref{eq:eq1}) but uses kernel ridge regressions instead of Gaussion processes:
\begin{align*}
\rho(x)&=\Phi_\rho(x)\boldsymbol{\beta}_\rho,\;\;\;\; \beta_\rho\sim\mathcal{N}(\hat{\boldsymbol{\beta}}_{\rho},\hat{\Sigma}_{\rho});\\
\delta(x)&=\Phi_\delta(x)\boldsymbol{\beta}_\delta,\;\;\;\; \beta_\delta\sim\mathcal{N}(\hat{\boldsymbol{\beta}}_{\delta},\hat{\Sigma}_{\delta}).
\end{align*}
This yields the $f_a$ of the form
\begin{align}
f_a&=\rho(x)f_p(x)+\delta(x)\\
&=\Phi_\rho(x)\boldsymbol{\beta}_{\rho}diag(f_p(x))+\Phi_\delta(x)\boldsymbol{\beta}_{\delta}.
\label{eq:fa_RAA}
\end{align}

\subsubsection{Ridge regression adjustment (RRA): learning \& prediction}
The calculating posterior distributions of $\boldsymbol{\beta}_\rho$ and $\boldsymbol{\beta}_\delta$ involves packing the feature matrices of training inputs, $\Phi_{X,\rho}$ and $\Phi_{X,\delta}$, together and applying the closed-form solution.
\begin{align*}
\boldsymbol{y}&=\Phi_{X,\rho}diag(f_p(X))\boldsymbol{\beta}_\rho+\Phi_{X,\delta}\boldsymbol{\beta}_\delta\\
&=\Phi_{X,\rho,f_p}\boldsymbol{\beta}_\rho+\Phi_{X,\delta}\boldsymbol{\beta}_\delta\\
&=\boldsymbol{\Phi}_{concat}\boldsymbol{\beta}_{concat}^T,
\end{align*}
where $\boldsymbol{\Phi}_{concat}=[\Phi_{X,\rho,f_p},\Phi_{X,\delta}]$ and $\boldsymbol{\beta}_{concat}=[\boldsymbol{\beta}_\rho,\boldsymbol{\beta}_\delta]$ are the concatenation along the column. The posterior mean $\hat{\boldsymbol{\beta}}_{concat}$ and variance $\hat{\Sigma}_{concat}$ can then be computed using Eq.\ref{eq:kernel_ridge_closed_form_solution} and \ref{eq:beta_dist}.

The predictive mean $\boldsymbol{\mu}_{x_*}$ at input location $x_*$ can be computed by substituting ${\boldsymbol{\beta}}_\rho$ and ${\boldsymbol{\beta}}_\delta$ in Eq.\ref{eq:fa_RAA} with $\hat{\boldsymbol{\beta}}_\rho$ and $\hat{\boldsymbol{\beta}}_\delta$. The predictive variance $\boldsymbol{var}_{x_*}$ can be computed similarly to Eq.\ref{eq:RRA_predictive_var} as
\begin{equation}
\boldsymbol{var}_{x_*} = [\Phi_\rho(x_*),\Phi_\delta(x_*)]^T\hat{\Sigma}_{concat}[\Phi_\rho(x_*),\Phi_\delta(x_*)].
\end{equation}
Alternatively, it is possible to draw a set of deterministic functions by sampling multiple $\boldsymbol{\beta}_{concat}$ from the posterior distribution and applying them to Eq.\ref{eq:fa_RAA}.

\subsubsection{RBF kernel approximation with random Fourier features}
The behaviour of kernel ridge regression is dictated by the kernel function. Here, we want RRA to have the behaviour of RBF kernel which can be approximated using random Fourier features \citep{Rahimi2007}. The RBF kernel with length scale $l$ is approximated as
\begin{equation}
\varphi_j(x) = cos(-\boldsymbol{\omega}_jx+b_j),
\end{equation}
where $\boldsymbol{\omega}_j\sim\mathcal{N}(\mathbf{0},\boldsymbol{I}/l)$, and $b_j\sim\text{Uniform}(0,2\pi)$. The feature vector of an input $x$ is the aggregation of random features: $\boldsymbol{\Phi}_x = \sqrt{2/D}[\varphi_1(x),...,\varphi_D(x)]$, where $D$ is the number of random features.

\subsection{Physics-derived model}
CKA and RRA requires a physics-derived model $f_p(x)$ that predicts the next state $\boldsymbol{s}_{t+1}$ from the current state $\boldsymbol{s}_t$ and action (or control input) $\boldsymbol{a}_t$. CKA and RRA are very flexible on how $f_p(x)$ is computed: $f_p(x)$ can be in different forms, e.g., explicit functions, black box models, or simulators, and does not need to be differentiable. In this work, we demonstrate two scenarios in which  $f_p(x)$ is computed using 1) a simulator whose underlying functions are unknown and 2) an ordinary differential equation (ODE). The simulator scenario is straightforward, as the prediction can be directly computed. The ODE scenario, which usually has an ODE in the form of $\dot{\boldsymbol{s}_t}=f_{\mbox{ODE}}(\boldsymbol{s_t,a_t})$, requires the integration, which can be computed numerically, over a certain time-step size $t_s$ with $\boldsymbol{s}_t$ and $\boldsymbol{a}_t$ as initial conditions. This is simply expressed as $f_p(\boldsymbol{s}_t,\boldsymbol{a}_t) = \int_{t=0}^{t=t_s}f_{\mbox{ODE}}(s_t,a_t)\,dt$. We assume that $\boldsymbol{a}_t$ is constant during the time step.


\subsection{CKA and RRA in reinforcement learning}
CKA and RRA can be used in reinforcement learning in different ways. Here, to present their generality and effectiveness, we demonstrate their uses in Dyna architecture \citep{Sutton1990}. In a basic format, a Dyna architecture treats the learnt model as a replicated environment with which the agent can interact and collect the experience tuples. The policy learning or updating mechanism itself remains unchanged.

The reinforcement learning algorithm of choice here is soft actor-critic (SAC) with double Q networks \cite{Haarnoja2019_1}. SAC is chosen because it requires less parameter tuning and has robust performances across different environments. Originally, SAC was an off-policy model-free reinforcement learning algorithm with a stochastic policy. The policy is optimised using experience tuples $(\textbf{s},\textbf{a},r,\textbf{s}_{t+1})$ stored in a replay buffer $\mathcal{D}$. In our Dyna architecture here, we use two replay buffers, $\mathcal{D}_R$ and $\mathcal{D}_M$, to separately store the real and model-based (the interaction with CKA) experience tuples. In the policy update, the tuples from $\mathcal{D}_R$ and $\mathcal{D}_M$ are mixed with the ratio of 1:10. The model-based interaction trajectories are generated using a particle approach: sample the next states from the predictive distribution and repeat.

\section{Experiments}

\subsection{Forester function}
The Forrester function \citep{Forrester2007} is a standard one-dimensional test task in multi-fidelity modelling \cite{Raissi16}. The task is to fit the \emph{true} function using the data from the \emph{true} and \emph{crude} functions:
\begin{align*}
true function:&\;\; f_{\mbox{true}}(x)=\frac{1}{4}(6x-2)^2sin(12x-4)\\
crude function:&\;\; f_p(x)=\frac{1}{4}(\frac{f_{\mbox{true}}(x)}{2}+10(x-\frac{1}{2})+5)
\end{align*}
The crude function can be considered as the physics model ($f_p(x)$).

We compare the behaviours of 7 models--zero-mean GP, Phy-mean GP, GP-bias, GP-scale, AR1 co-kriging, CKA, and RRA--on this task with 8 observations randomly sampled from the true function. The Phy-mean GP uses $f_p$ as the mean function. The GP-bias uses a GP to learn only bias term in the adjustment function ($\delta(x)$ in Eq.\ref{eq:eq1}). The Phy-mean GP and GP-bias are, in fact, equivalent. GP-scale learns only the scaling term ($\rho(x)$ in Eq.\ref{eq:eq1}). The AR1 model is trained with 40 $f_p$-generated data points, uniformly distributed over $x\in[-0.6,1.0]$. All models use RBF kernel.

\begin{figure}[h!]
    \begin{center}
    \includegraphics[width=\columnwidth]{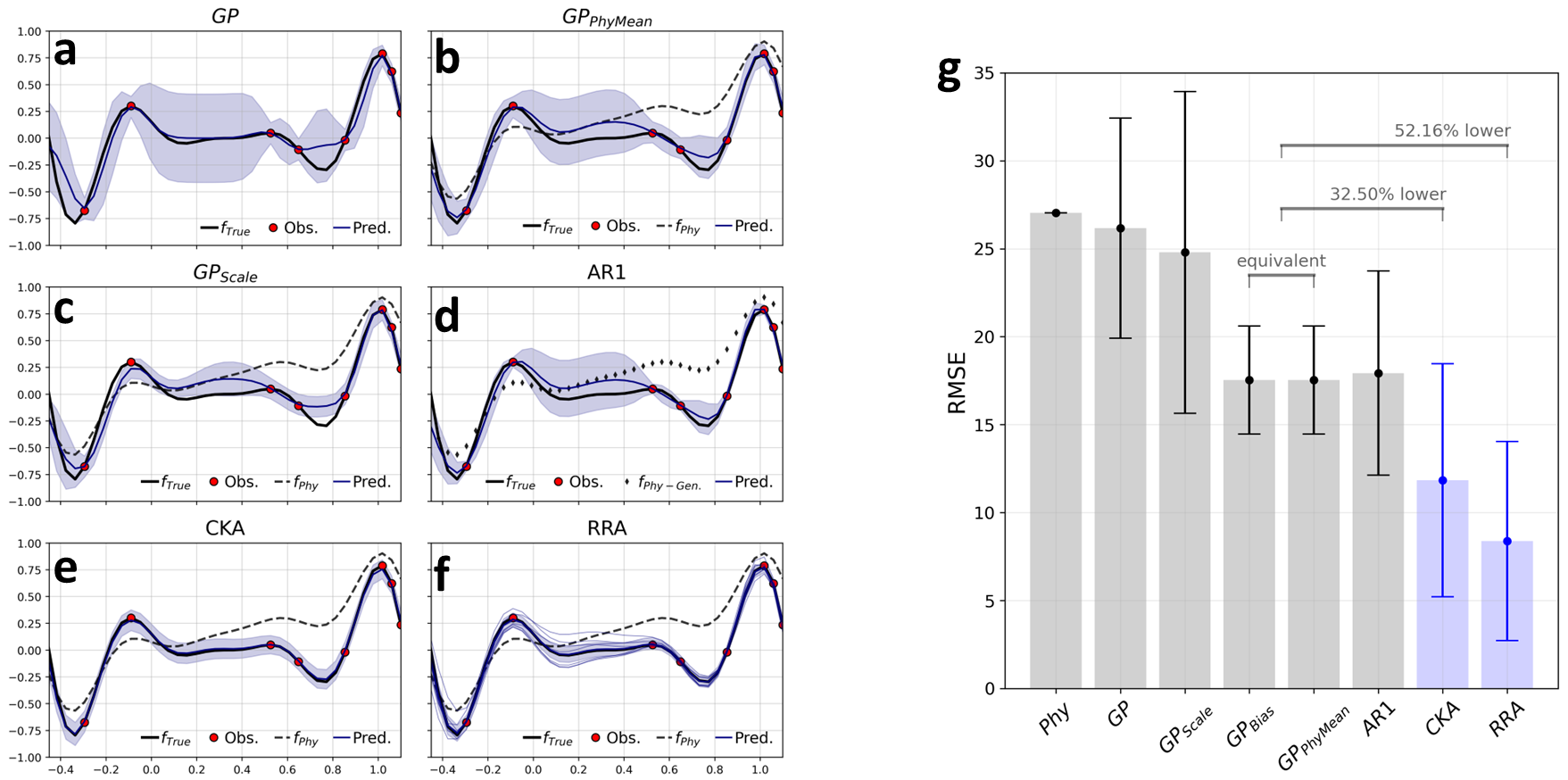}
    \caption{Example Forester function fitting of (a) zero-mean GP, (b) Phy-mean GP, (c) GP-scale, (d) AR1 co-kriging, (e) CKA, and (f) RRA. The dash and solid lines represent the crude and true functions, respectively. The red dots are the sample observations drawn from the true function. The navy lines and shades represent the predictive mean and standard deviation of the models. The thin navy lines in (f) are sampled deterministic functions. (g) The RMSEs between the predictive means and the true function of all models over 100 runs. The error bars represents the standard deviation.}
    \vspace{-5mm}
    \label{fig:forrester_fitting}
    \end{center}
\end{figure}

Fig.\ref{fig:forrester_fitting}a-f show example fitting of all models except GP-bias because it is equivalent to Phy-mean GP. All models provides good fitting at the locations of the observation. Zero-mean GP does not perform well at other locations (Fig.\ref{fig:forrester_fitting}a). Phy-mean GP, GP-scale, and AR1 behave similarly (Fig.\ref{fig:forrester_fitting}b-d). Their predictive means become closer to $f_p(x)$ and variances become larger with the locations farther from the observations. CKA and RRA have the predictive means that almost perfectly match the true function (Fig.\ref{fig:forrester_fitting}e and f). Fig.\ref{fig:forrester_fitting}g shows the RMSEs, computed from 40 points over $x\in[-0.6,1.0]$, between the predictive means and the true function over 100 runs with different randomly-sampled observations. The left most column shows that the $f_p(x)$ is relatively inaccurate. Zero-mean GP's RMSE is slightly better than that of $f_p(x)$. GP-scale performs slightly better than zero-mean GP but has larger standard deviation.  Phy-mean GP, GP-bias (equivalent to Phy-mean GP), GP-scale, and AR1 co-kriging have better RMSEs than GP-scale. AR1 has the biggest standard deviation amongst the three. CKA is the second best performer; its RMSE is roughly 30\% smaller that that of Phy-mean GP. RRA is the best performer, with roughly 50\% smaller RMSE compared to Phy-mean GP.

\subsection{Invert Pendulum}
\begin{figure}[h!]
    \begin{center}
    \includegraphics[width=\columnwidth]{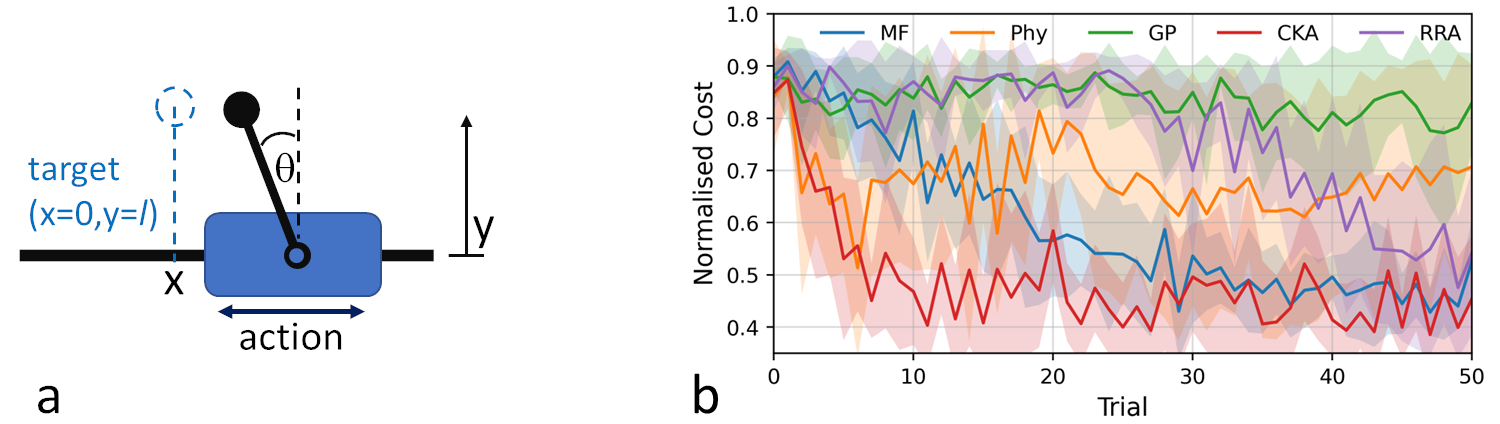}
    \caption{(a) The illustration of the invert pendulum system. (b) The learning curves of scenarios. The solid lines and shades represent the mean and standard deviation of 10 runs.}
    \vspace{-5mm}
    \label{fig:invert_pendulum}
    \end{center}
\end{figure}

Inverted pendulum is a standard benchmark system in control and reinforcement learning. This system has a cart that moves along x-axis and a hinged pendulum attached to the cart (Fig.\ref{fig:invert_pendulum}a). The task is to apply force $a\in[-10,10]\:N$, that pushes the cart along x-axis, in order to swing and balance the pendulum at an upright position. In this experiment, we follows the setups used in \citep{Deisenroth2011, Gal2016}  in which the state vector comprises the cart position, pendulum angle, and their
time derivatives $\boldsymbol{s}=[x,\theta,\dot{x},\dot{\theta}]^T$. The differential equation of motion is
\begin{equation*}
\frac{d\boldsymbol{s}}{dt} = [\dot{x},\dot{\theta},\frac{2m_pl\dot{\theta}^2s+3m_pgsc+4a-4b\dot{x}}{4(m_c+m_p)-3m_pc^2},\frac{-3m_pl\dot{\theta}^2s c-4(m_c+m_p)gs-6(a-b\dot{x})c}{4l(m_c+m_p)-3m_plc^2}]^T,
\end{equation*}
where $m_c$ and $m_p$ are the mass of the cart and the pendulum, respectively; $l$ is the pendulum's length; $b$ is damping coefficient; and $s$ and $c$ are the shorthand for $sin\theta$ and $cos\theta$, respectively.

The real system has $m_c=m_p=0.5kg$, $l=0.6m$, and $b=0.1Ns/m$. The Physics model $f_p(x)$ has the parameters sampled from uniform distributions as follows: $m_c\sim U(0.4,0.6)$, $m_p\sim U(0.5,0.7)$, and $l\sim U(0.5,0.7)$. The damping is not included in the $f_p(x)$, i.e., $b=0Ns/m$. The cost function is $1-exp(-2d^2)$, where $d$ is the Euclidean distance between the pendulum's end point and the goal $[x=0,y=l]$. One trial comprises 25 steps with the time-step size of $0.1s$. Each trial start with the initial state $\boldsymbol{s}_0\sim\mathcal{N}(\boldsymbol{\mu},\Sigma)$, where $\boldsymbol{\mu}=[0,\pi,0,0]$ and $\Sigma^\frac{1}{2}=diag([0.2,0.2,0.2,0.2])$.

We apply soft-actor critic (SAC) in 5 scenarios: model-free (MF), Dyna with Physics model (solely $f_p(x)$ without adjustment functions), Dyna with GP, CKA, and RRA. In each scenario, the training is 50 trials long with 10 repetitions. For Dyna scenarios, 20 model-based training trials are carried out after each real trial. The model-based training is not performed after 25 real trials.

Fig.\ref{fig:invert_pendulum}b shows the learning curves plotted as normalised episode cost against trials. The normalised cost below 0.5 means the RL agent can swing up and balance the pendulum with some errors. From Fig.\ref{fig:invert_pendulum}, model-free SAC reaches the 0.5 cost level in 30 trials and occasionally touches 0.4 level within 50 trials. Dyna with only physics model reaches the 0.5 level in 6 trials but gradually loses the performance as the training progresses. It is also unable to learn the task after switching to the pure model-free training in the last 25 trials. Dyna-GP is unable to learn the task. Dyna-CKA is the best performer; it reaches 0.5 level within 6 trials, periodically touches 0.4 level after 12 trials, and can maintain good performance throughout 50 trials. Surprisingly, Dyna-RRA cannot learn the task during the model-based training but can catch Dyna-CKA in the last 25 trials when switched to the pure model-free training.


\subsection{Planar arm control via muscle activation}
\begin{figure}[htb!]
    \begin{center}
    \includegraphics[width=1.0\columnwidth]{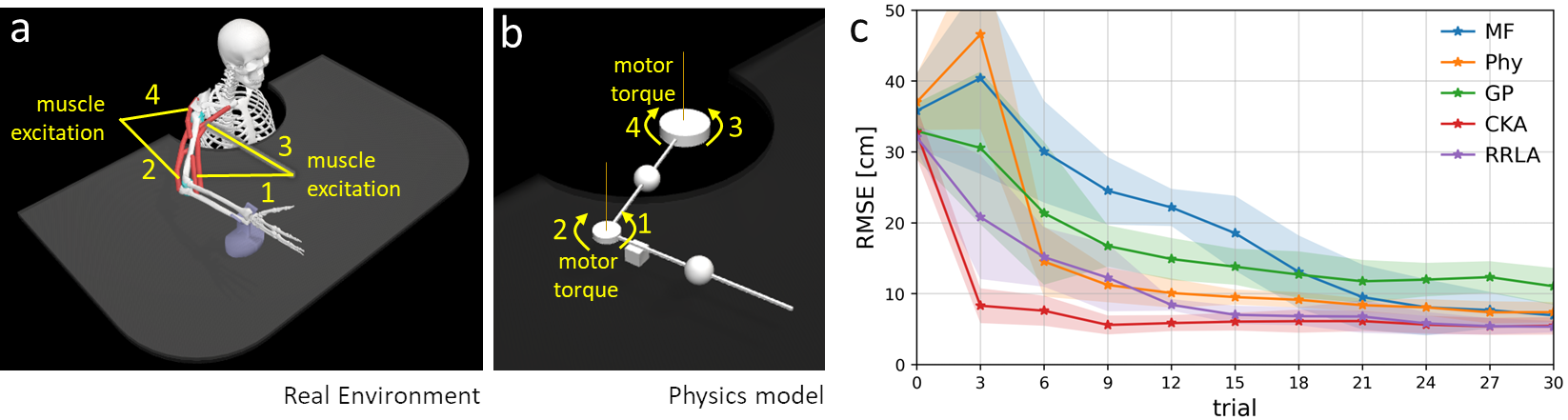}
    \caption{Human arm control task. (a) the real system of human with 6 muscles in OpenSim. (b) a two-link arm with 3 torque actuators as a physics-derived model ($f_p$) for CKA.}
    \vspace{-2mm}
    \label{fig:arm}
    \end{center}
\end{figure}

In this task, we demonstrate the effectiveness of CKA and RRA in controlling a neuromechanical system: a 2-DoF human arm moving on a table (Fig.\ref{fig:arm}a). The arm has 6 muscles; 4 muscles which are \emph{biceps}, \emph{triceps}, and \emph{deltoid anterior} are excited (controlled) through action $a_1,a_2,a_3,a_4$, respectively. This is a complex dynamical system for several reasons. Firstly, the relationship between force production and muscle stimulation is nonlinear and depends on many factors, such as muscle fibre length, contraction speed, and the muscle's activation delay. Secondly, the muscle applies force across a joint (or two joints for some muscles) to produce the torque, which varies non-linearly with the joint angle. Additionally, the non-stimulated muscles have passive force due to their elasticity.

In this experiment, we assume that we do not know the exact dynamics of the arm. However, we know that its behaviour is similar to a 2-DoF robot arm (Fig.\ref{fig:arm}b) whose dynamics is known and can be, therefore, used as physics-derived model ($f_p(x)$). The robot arm has the same length as the arm and a 10\% error in mass. The actuation is simplified to torque actuators. The actions $[a_1,a_2,a_3,a_4]^T$ applied to the robot arm directly induce the torque on the joints. Small torque damping is applied to both joints to prevent unrealistic angular velocities. This robot arm model is also created in OpenSim to demonstrate that $f_p(x)$ is computed using a simulator. It is pertinent to mention that, in the case of high-dimensional robotic tasks where ODEs are computationally expensive and impractical to obtain, simulators like OpenSim or Mujoco can be used effectively.

On the reinforcement learning side, the task is to apply muscle excitation to achieve the target elbow and shoulder angles. The reinforcement learning state vector comprises shoulder and elbow angles, their time derivative, and the target $\boldsymbol{s}=[\theta_s; \theta_e; \dot{\theta_s}; \dot{\theta_e}; \theta_{\mbox{s-tar}}; \theta_{\mbox{e-tar}}]$. The cost (negative reward) function is simply the square error between the target and actual angles, computed as $cost(t) = RMSE([\theta_{\mbox{s-tar}};\theta_{\mbox{e-tar}}], [\theta_{\mbox{s,t}};\theta_{\mbox{e,t}}])$.

Similar to the invert pendulum experiment, we explore also those 5 scenarios with similar training setup. Each trial starts with the initial state $\boldsymbol{s}_0\sim\mathcal{N}(\mu,\Sigma)$, where $\mu=[35\pi/180,65\pi/180,0,0]$ and $\Sigma^\frac{1}{2}=diag([5\pi/180,5\pi/180,0,0])$. The target angles $[\theta_{\mbox{s-tar}},\theta_{\mbox{e-tar}}]$ are sampled from a uniform distribution with the interval $[0,100\pi/180]$. Each trial has 15 steps with the time-step size of $0.2s$. There are 20 model-based trials after each real trial. Every 3 trials, a performance evaluation is carried out on reaching tasks with 100 targets, the permutation of the target angle vector where $\theta_{\mbox{s-tar}}$ and $\theta_{\mbox{e-tar}}$ are the set of $[0,10\pi/180,...,100\pi/180]$.

Fig.\ref{fig:arm}c shows the performance evaluations along the training over 10 runs in terms of RMSE between the hand and the target in centimetres. At the end of the training, the model-free (MF) scenario has the RMSE around 8 cm. Dyna-GP has steeper learning curve than MF but its performance stagnates at the RMSE above 10 cm. Incorporating the physics model into the training provides substantial boost to the learning. Dyna-Phy is able to reach 10 cm RMSE within 12 trials but improves slowly thereafter. Dyna-CKA is, again, the best performer; it reaches the 8 cm level within 3 trials and continues to improve thereafter, achieving the RMSE around 6 cm at the end of the training. Dyna-CKA also has the lowest variances. Dyna-RRA is the second best performer; it learns as fast as Dyna-Phy but continues to improve and catches Dyna-CKA within 24 trials.

\begin{figure}[h!]
    \begin{center}
    \includegraphics[width=1.0\columnwidth]{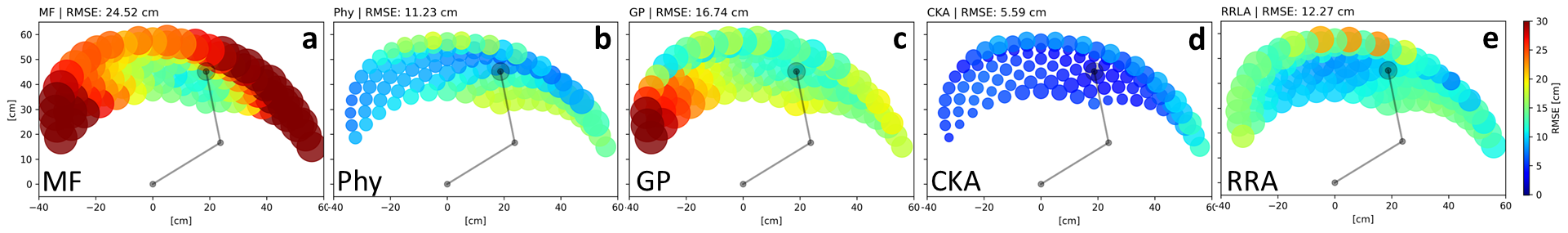}
    \caption{Performance evaluation after trial 9. The colour dots show 100-target average reaching error over 10 runs of (a) model-free, (b) Dyna-Phy, (c) Dyna-GP, (d) Dyna-CKA, and (e) Dyna-RRA scenarios. The dots' sizes correspond to the standard deviation (the larger, the higher). The grey links represent the initial arm positions.}
    \vspace{-2mm}
    \label{fig:100target}
    \end{center}
\end{figure}

Fig.\ref{fig:100target} shows 100-target reaching error from the performance evaluation after trial 9. The MF scenario still has very large error over the entire workspace at this stage (Fig.\ref{fig:100target}a). It performs relatively well around the locations of the initial position. Dyna-GP has better overall performance than MF (Fig.\ref{fig:100target}c). However, it still fails to reach the targets in far regions. Substantial improvements over the entire workspace appear in the scenario that incorporate the physics model. Amongst those, Dyna-CKA is outstanding; it achieves low error and variance at nearly all targets, including those in the far regions.

\section{Conclusions and Discussions}

We present two simple and flexible approach to incorporate physics knowledge about system dynamics into probabilistic machine learning models. Our approach, co-kriging adjustment (CKA) and ridge regression adjustment (RRA), is based on the idea of learning the probabilistic adjustment functions to make the outputs of physics-derived model match the observation. We present the fitting behaviours of CKA and RRA in learning a benchmark function, highlighting the differences between CKA, RRA and the existing methods. Next, we demonstrate an application of CKA and RRA in reinforcement learning with a Dyna architecture and show that CKA contributes to a significant improvement in data efficiency, even when the physics models of the environment do not fully capture the dynamics of the real systems.

There are several ways to expand the use of CKA. One potential approach is to combine CKA with information-driven exploration to efficiently gather the data over the entire state-action space. The model can then be used to devise a target-directed policy that can operates over the entire space. Secondly, CKA is compatible with probabilistic policy search techniques such as PILCO \citep{Deisenroth2011} or PIPPS \citep{Parmas2018}. Combining CKA with such practices could substantially improve data efficiency, bringing reinforcement learning closer to practical usage in the real worlds, especially in learning the control or manipulation of complex systems.


{
\small

\bibliography{bib}

\begin{thebibliography}{24}
\providecommand{\natexlab}[1]{#1}
\providecommand{\url}[1]{\texttt{#1}}
\expandafter\ifx\csname urlstyle\endcsname\relax
  \providecommand{\doi}[1]{doi: #1}\else
  \providecommand{\doi}{doi: \begingroup \urlstyle{rm}\Url}\fi

\bibitem[Andersson et~al.(2015)]{andersson2015model}
O.~Andersson et~al.
\newblock Model-based reinforcement learning in continuous environments using real-time constrained optimization.
\newblock In \emph{29th AAAI}, 2015.

\bibitem[Brevault et~al.(2020)Brevault, Balesdent, and Hebbal]{BREVAULT2020}
L.~Brevault, M.~Balesdent, and A.~Hebbal.
\newblock Overview of gaussian process based multi-fidelity techniques with variable relationship between fidelities, application to aerospace systems.
\newblock \emph{Aerospace Science and Technology}, 107, 2020.

\bibitem[Cen et~al.(2011)Cen, Wei, and Jiang]{Cen2011}
Z.~Cen, J.~Wei, and R.~Jiang.
\newblock A grey-box neural network based identification model for nonlinear dynamic systems.
\newblock \emph{Proc. of 4th Intl. Workshop on Adv Comp. Intel. (IWACI) 2011}, pages 300--307, 2011.

\bibitem[Deisenroth and Rasmussen(2011)]{Deisenroth2011}
M.~P. Deisenroth and C.~E. Rasmussen.
\newblock Pilco: A model-based and data-efficient approach to policy search.
\newblock In \emph{28th Intl. Conf. Machine Learning (ICML)}, pages 465--472, 2011.

\bibitem[Forrester et~al.(2007)Forrester, Sóbester, and Keane]{Forrester2007}
A.~I. Forrester, A.~Sóbester, and A.~J. Keane.
\newblock Multi-fidelity optimization via surrogate modelling.
\newblock \emph{Proceedings of the Royal Society A: Mathematical, Physical and Engineering Sciences}, 463:\penalty0 3251--3269, 12 2007.

\bibitem[Gal et~al.(2016)Gal, Mcallister, and Rasmussen]{Gal2016}
Y.~Gal, R.~T. Mcallister, and C.~E. Rasmussen.
\newblock Improving pilco with bayesian neural network dynamics models.
\newblock \emph{Data-Efficient Machine Learning Workshop, ICML}, pages 1--7, 2016.

\bibitem[Gratiet and Garnier(2014)]{Gratiet2014}
L.~L. Gratiet and J.~Garnier.
\newblock Recursive co-kriging model for design of computer experiments with multiple levels of fidelity.
\newblock \emph{Intl. J. Uncertainty Quantification}, 4:\penalty0 365--386, 2014.

\bibitem[Haarnoja et~al.(2019)]{Haarnoja2019_1}
T.~Haarnoja et~al.
\newblock Soft actor-critic algorithms and applications.
\newblock \emph{arXiv:1812.05905v2 [cs.LG]}, 2019.

\bibitem[Kennedy and O'hagan(2000)]{Kennedy2000}
B.~M.~C. Kennedy and A.~O'hagan.
\newblock Predicting the output from a complex computer code when fast approximations are available.
\newblock \emph{Biometrika}, 87:\penalty0 1--13, 2000.

\bibitem[Kormushev et~al.(2013)Kormushev, Calinon, and Caldwell]{Kormushev2013}
P.~Kormushev, S.~Calinon, and D.~G. Caldwell.
\newblock Reinforcement learning in robotics: Applications and real-world challenges.
\newblock \emph{Robotics}, 2\penalty0 (3), 2013.

\bibitem[Nagabandi et~al.(2017)]{Nagabandi2017}
A.~Nagabandi et~al.
\newblock Neural network dynamics for model-based deep reinforcement learning with model-free fine-tuning.
\newblock In \emph{Deep Reinforcement Learning Symposium, NIPS}, 2017.

\bibitem[Nguyen and La(2019)]{Nguyen2019}
H.~Nguyen and H.~La.
\newblock Review of deep reinforcement learning for robot manipulation.
\newblock In \emph{3rd IEEE Intl. Conf. Robotic Computing (IRC)}, pages 590--595, 2019.

\bibitem[Parmas et~al.(2018)]{Parmas2018}
P.~Parmas et~al.
\newblock Pipps: Flexible model-based policy search robust to the curse of chaos.
\newblock In \emph{Intl. Conf. Machine Learning (ICML)}, 2018.

\bibitem[Rahimi and Recht(2007)]{Rahimi2007}
A.~Rahimi and B.~Recht.
\newblock Random features for large-scale kernel machines.
\newblock In \emph{NIPS}, 2007.

\bibitem[Raissi and Karniadakis(2016)]{Raissi16}
M.~Raissi and G.~E. Karniadakis.
\newblock Deep multi-fidelity gaussian processes.
\newblock \emph{arXiv:1604.07484v1 [cs.LG]}, 2016.

\bibitem[Raissi et~al.(2019)Raissi, Perdikaris, and Karniadakis]{RAISSI2019}
M.~Raissi, P.~Perdikaris, and G.~Karniadakis.
\newblock Physics-informed neural networks: A deep learning framework for solving forward and inverse problems involving nonlinear partial differential equations.
\newblock \emph{J. Computational Physics}, 378:\penalty0 686--707, 2019.

\bibitem[Ramesh and Ravindran(2022)]{ramesh2022physics}
A.~Ramesh and B.~Ravindran.
\newblock Physics-informed model-based reinforcement learning.
\newblock \emph{arXiv preprint arXiv:2212.02179}, 2022.

\bibitem[Sun et~al.(2022)]{sun2022physics}
W.~Sun et~al.
\newblock Physics-informed recurrent neural networks for soft pneumatic actuators.
\newblock \emph{IEEE Robotics and Automation Letters}, 7\penalty0 (3):\penalty0 6862--6869, 2022.

\bibitem[Sutton(1991)]{Sutton1990}
R.~S. Sutton.
\newblock Dyna, an integrated architectures for learning, planning, and reacting based on approximating dynamic programming.
\newblock \emph{SIGART Bulletin}, 2\penalty0 (4):\penalty0 160--163, 1991.

\bibitem[Taneja et~al.(2022)]{taneja2022feature}
K.~Taneja et~al.
\newblock A feature-encoded physics-informed parameter identification neural network for musculoskeletal systems.
\newblock \emph{J. Biomechanical Engineering}, 144\penalty0 (12), 2022.

\bibitem[Tulleken(1993)]{Tulleken1993}
H.~J. Tulleken.
\newblock Grey-box modelling and identification using physical knowledge and bayesian techniques.
\newblock \emph{Automatica}, 29:\penalty0 285--308, 1993.

\bibitem[Wannawas et~al.(2021)Wannawas, Subramanian, and Faisal]{Wannawas2021}
N.~Wannawas, M.~Subramanian, and A.~A. Faisal.
\newblock Neuromechanics-based deep reinforcement learning of neurostimulation control in fes cycling.
\newblock In \emph{Intl. IEEE/EMBS Conf. on Neural Engineering (NER)}, volume 2021-May, pages 381--384, 2021.

\bibitem[Wannawas et~al.(2022)Wannawas, Shafti, and Faisal]{Wannawas2022}
N.~Wannawas, A.~Shafti, and A.~A. Faisal.
\newblock Neuromuscular reinforcement learning to actuate human limbs through fes.
\newblock In \emph{IFESS}, 2022.

\bibitem[Zhao et~al.(2020)Zhao, Queralta, and Westerlund]{Zhao2021}
W.~Zhao, J.~P. Queralta, and T.~Westerlund.
\newblock Sim-to-real transfer in deep reinforcement learning for robotics: a survey.
\newblock In \emph{IEEE Symposium Series on Computational Intelligence (SSCI)}, 2020.

\end{thebibliography}

}

\end{document}